# Anomaly Detection Using the Knowledge-based Temporal Abstraction Method

Asaf Shabtai
Dept. of Software and Information Systems Engineering,
Ben-Gurion University of the Negev, Beer-Sheva, Israel
shabtaia@bgu.ac.il

*Abstract*— **The rapid growth in stored time-oriented data necessitates the development of new methods for handling, processing, and interpreting large amounts of temporal data. One important example of such processing is detecting anomalies in time-oriented data. The Knowledge-Based Temporal Abstraction method was previously proposed for intelligent interpretation of temporal data based on predefined domain knowledge. In this study we propose a framework that integrates the KBTA method with a temporal pattern mining process for anomaly detection. According to the proposed method a temporal pattern mining process is applied on a dataset of basic temporal abstraction database in order to extract patterns representing normal behavior. These patterns are then analyzed in order to identify abnormal time periods characterized by a significantly small number of normal patterns. The proposed approach was demonstrated using a dataset collected from a real server.**

*Keywords*— *Anomaly detection, Knowledge-based Temporal Abstraction, KBTA, Temporal Data Mining.*

## I. Introduction

Today, data storage capabilities as well as computational power are rapidly increasing. On the one hand, this improvement makes it possible to generate and store a great amount of temporal data for future query, analysis and discovery of new knowledge. On the other hand, systems and experts are encountering new problems in processing this increased amount of data. The rapid growth in stored time-oriented data necessitates the development of new methods for handling, processing, and interpreting large amounts of temporal data. One approach is to use an automatic summarization process based on predefined knowledge, such the Knowledge-Based Temporal-Abstraction (KBTA) method [1]. This method enables one to summarize and reduce the amount of raw data by creating higher level interpretations based on predefined domain knowledge.

*Temporal-Abstraction* (TA) is the task of integrating raw time-stamped data and knowledge in order to extract and summarize meaningful, context-sensitive, conclusions and interpretations (i.e., temporal abstractions) from these data. Shahar's *Knowledge-Based Temporal-Abstraction* (KBTA) method [1] is a computational framework for supporting the TA task by decomposing it into several computational sub-tasks. The KBTA method provides automated means for deriving context-specific temporal abstractions from raw, time-oriented data by using a domain-specific knowledge base (e.g., a set of instances of a specialized security ontology for abstracting meaningful patterns from time-oriented security data). In general, the KBTA method is defined as follows. The input includes a set of raw, time-stamped, concepts (e.g., a number of FTP connections at each time-point) and events (e.g., keyboard or mouse activity) that create the necessary interpretive context (e.g., "no user activity" when the mouse and keyboard are not being used). The output includes a set of interval-based, context-specific concepts at the same or at a higher level of abstraction and their respective values. A typical example of security-based output would be a period of two hours of a high number of FTP connections while no user activity was detected. Such an output may indicate that a Trojan horse is installed in the user computer and is leaking information via FTP connections.

A knowledge engineer defines the domain knowledge relevant to the TA task by using a temporal-abstraction ontology that includes five KBTA entities (concepts, events, contexts, abstractions and patterns), their properties, and the relations among them. Five inference mechanisms -- temporal context formation, contemporaneous abstraction, temporal inference, temporal interpolation and temporal pattern matching – that essentially operate in parallel, are then applied to derive the high level abstractions from the raw data [2]. The KBTA method was originally proposed for assisting physicians in typical tasks such as therapy, quality assessment monitoring and clinical trial analysis in medicine [2][3]; and later was extended to intrusion detection in computer-network security [4][5]; fraud detection and decision-making in the financial domains [6]; and analysis of information and decision-making in military intelligence. This is done by defining the interesting, and meaningful patterns.

This research presents the adaptation and application of the KBTA method with a time interval mining process for anomaly detection. Generally, the KBTA can be employed for anomaly detection by applying the two main approaches: (1) specifying *abnormal* pattern in the KBTA domain ontology; and (2) specifying temporal patterns representing normal behavior.

This research focuses on the second approach; i.e., mining, from abstract data, temporal patterns which represent the normal behavior of a subject and identifying time intervals in which only few these normal patterns exist. A preliminary evaluation of this approach is presented. We perform a chronological evaluation, using data collected from a real server, in which we derive normal patterns during a period of one week and use the derived patterns for detecting abnormal time intervals on the following week. Abnormal intervals are defined as time intervals in which a relatively small number of normal behavioral patterns is identified.

## II. KNOWLWDGE-BASED TEMPORAL ABSTRACTION

The *knowledge based temporal abstraction* (*KBTA*) method receives as input raw measured data and external events and returns a set of interval-based, context-specific concepts at the same or at higher level of abstraction and their respective values. Abstractions can include one of four types: state, gradient, rate, and temporal patterns. A state abstraction is an abstraction of the values of one or more contemporaneous concepts to a "state-describing" set of values; for example, the state of bytes sent/sec can be 'low', 'medium' or 'high' and is abstracted from the primitive concept, "number of bytes sent/sec". The gradient abstraction determines the direction of the change of values in a measured concept (e.g., 'increasing' number of fails connections). The rate abstraction classifies the amplitude of a rate of change of a selected concept (e.g., rapidly changing number of modified exe files). Finally, temporal patterns are defined as a complex set of value and time constraints defined over a set of concepts (both primitive and abstract), events, and contexts. Temporal patterns can be linear (a set of value and temporal relationship constraints that hold among several raw and/or abstract concepts) or repeating (two or more instances of linear or periodic patterns, amongst which certain periodic constraints hold). The appropriate temporal contexts for the abstraction process are induced recursively from events and from derived abstractions [7] (e.g., the event 'connecting to an unprotected Wi-Fi' will induce a 'Connected to unprotected Wi-Fi' context which will be used for the abstraction of other concepts. Abstract concept intervals are interpolated from time-stamped concepts or from intervals, using a temporal-interpolation model [8], into a longer time interval. The related theory was originally introduced in the medical field [2], but evaluated on other domains as well [5][9]. Figure 1 presents an example of the abstraction process. A primitive parameter 'the number of TCP connection failures' of a machine is abstracted into the TCP connection failure state abstraction' (four time intervals with the values normal, normal, high, and very high); in parallel, the raw number of TCP connection failures is abstracted into a gradient abstraction (increasing number of failed connections) and interpolated across the whole period.

## III. ANOMALY DETECTION USING THE KBTA METHOD

The KBTA method can be used for anomaly detection by applying the following approaches: (1) specifying abnormal patterns as part of the domain ontology; and (2) specifying normal patterns in the domain ontology and looking for time periods having significantly less normal patterns.

### A. Known Abnormal Patterns

In this approach the domain expert feeds the domain ontology with known abnormal patterns (e.g., patterns describing malicious behavior of a user, process or machine).

#### 1) Manual pattern mining

Anomalous temporal patterns can be identified by performing manual pattern mining. The domain expert can use a visual exploration tool [11] in order to examine both raw and abstracted data. The assumption here is that the domain expert has some knowledge and an indication of an abnormal event and the goal is to manually extract temporal patterns that leads or represents the event. The visual exploration tool enables the domain expert to examine the data in ways that can reveal new correlations or insights, thus acquiring new patterns of anomalies which are more complex than a person can perceive without using the tool.

#### 2) Temporal pattern mining

Similar to the previous approach, the assumption here is that an indication of a known anomalous event is provided. A temporal pattern mining algorithm (such as the algorithm introduced in the following section) is applied on the abstracted data that precedes the anomalous event and can extract significant temporal patterns. Such significant patterns may represent the anomalous event or represent the behavior that precedes such an event. The challenge in this approach is to identify these significant patterns. It is very hard to identify these patterns based on a single anomaly event instance. This approach requires several anomalous event instances that belong to the same category (e.g., database server crash, application server virus alert) in order to be identified as anomalous patterns.

### B. Normal Behavior Detection

In the normal behavior detection approach the temporal pattern mining algorithm is applied on the abstracted data to extract significant temporal patterns. In this case significant patterns represent, for example, the normal behavior of the server. An example of such a significant pattern can be a recurring pattern over time on a single server or a pattern recurring over multiple servers. The significant patterns found are being inserted to the knowledge base as desirable patterns. Several techniques can then be used to detect abnormal behavior; an example of a naïve technique is to use the skyline abstraction (presented in Section V). The skyline abstraction discretisize the number of normal behavior patterns occurring over each time interval. Abnormal behavior would be reflected in a low number of normal behavior patterns. The underlying assumption of this approach is that highly recurring patterns represent the normal behavior and therefore, anomalous behavior will be characterized by having significantly less normal patterns. The normal patterns are added to the domain ontology, the KBTA computational engine will extract these patterns from the data, and an alert will be issued upon the detection of a low number of normal patterns.

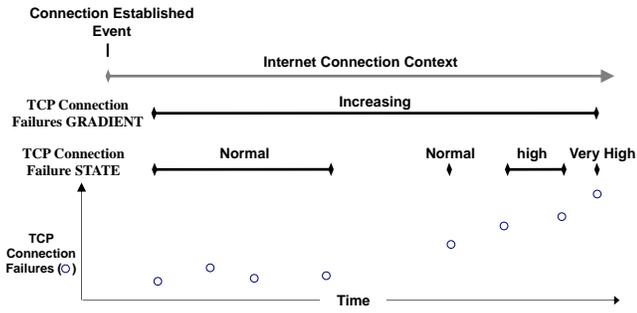

Fig. 1. Example of temporal abstraction in the network security domain.

## IV. INTERVAL MINING ALGORITHM

The temporal pattern mining algorithm that is used in this study is based on the algorithm proposed by Kam and Fu [11] with the following modifications: the addition of horizontal support, temporal coverage support, and the discretization of the duration of interval. While the algorithm presented in [11] considers Allen's 13 temporal relations [12], in our implementation we aggregate these relations into three primitive temporal relations: *before* (which includes also the *meets* relation), *overlaps*, and *during* (which includes *equal*, *starts* and *finishes* relations).

### A. Definitions

The mining algorithm operates on a given set of facts F{A,B,C,D…}. In our case, a fact is defined as a triplet: <concept name, value, duration> where, **concept name** is the name of a concept from the ontology, **concept value** is a valid value of the concept, and **concept duration** is a discretized value of the interval duration (into Short, Medium and Long). An example of a fact can be <BytesReceivedSec_STATE, HIGH, LONG> which refers to a long time interval of high number of received bytes per second.

**Definition 1**: A temporal pattern is defined recursively as follows: Let $f \in F$ be a single fact type; $f$ is a temporal pattern, also called an **atomic pattern**. Let $X$ and $Y$ be two temporal patterns and *rel* is one of the optional temporal relations, then ($X$ rel $Y$) is called a **composite temporal pattern**.

**Definition 2**: The size of temporal pattern is the number of atomic patterns in the temporal pattern.

**Definition 3:** k-item is defined as {{$A_1, A_2,…, A_k$} , {$rel_1, rel_1,…, rel_{k-1}$}, $P$} where $A_1, …, A_k$ are the facts in the $k$-item, $rel_1, …, rel_{k-1}$ are the relations in the $k$-item and $P$ is temporal pattern in terms of the facts {$A_1, …, A_k$} and the relations {$rel_1, …, rel_{k-1}$ }, $k \geq 1$. For example, the following is a 2-item: {{$A,B$)},{$rel_1$="overlaps"}, $P$=($A$ overlaps $B$)}.

We strict our interest to temporal patterns of the form ((…($A_1$ $rel_1$ $A_2$ ) $rel_2$ $A_3$)… $rel_{k-1}$ $A_k$). These are called the **A1 temporal patterns** [11].

### B. The Algorithm

The algorithm presented in this section is divided into two main steps. In the first step the data is transforming into an item-list representation. In the second step the A1 temporal patterns are being mined.

*[1] Step I: transforming the database into an item-list representation*

Consider a servers database shown in Figure 2. Each tuple contains subject id (i.e., server id), fact or an item (for example fact $A$ can be long time interval of high CPU usage state == <CPU_Usage_STATE, High, Long>, and start\end time of the fact. In the first step, we transform the database into an **item-list** database where each fact is associated with a list of tuples <subject, start time, end time>.

*[2] Step II: mining A1 temporal patterns*

The mining process takes multiple passes over the data. In each iteration, we start with a seed set of large-items which was found in the previous iterations. First, we use the seed set for generating new potentially large items, called candidates, $C_k$, by adding one atomic item to elements in the seed set (see example in Figure 3). The same atomic item can be added with different temporal relation (i.e., before, during, overlaps).

**Definition 4:** Horizontal support is the number of instances a candidate $C_k$ has for a single subject (e.g., server) data.

**Definition 5:** Vertical support is the percentage of subjects that supports a candidate $C_k$ which also satisfies the horizontal support.

**Definition 6:** Temporal coverage support is a modification of the horizontal support. The temporal coverage is the sum of time intervals of all instances, which belong to a specific subject, of candidate $C_k$. A candidate $C_k$ satisfies the temporal coverage supports if its temporal coverage is equal or greater than the temporal coverage support threshold.

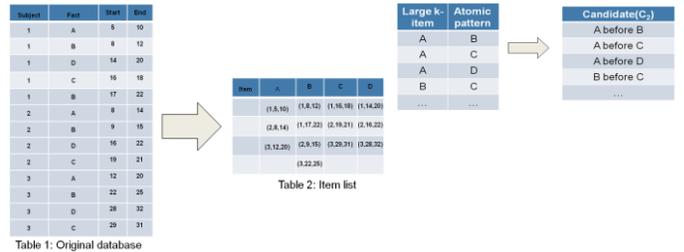

Fig. 2. DB to Item-list transformation.    Fig. 3. 2-candidate generation.

In the second phase (Fig. 4), for each candidate we examine the $L_{k-1}$ and $L_1$ item-lists and determine the temporal relations between the composite pattern and the atomic pattern that have sufficient vertical support as well as horizontal/temporal coverage support. The algorithm terminates when it cannot find any large $k$-items after the end of the current pass.

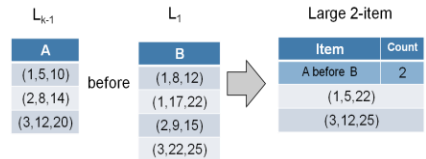

Fig. 4. Large 2-item generation.

## C. Mining Modes

One of the issues which affects the output of the mining process relates to cases in which the same element is shared among two or more instances of the same pattern. These cases can be handled in several ways from generating ALL possible combinations to generating specific patterns. We refer to these options as mining modes.

Consider, for example, facts A, B and their instances A1(1,2), A2(3,4), B1(5,6), B2(7,8) (Figure 5). For A1, there are two possible instances for the candidate (A before B): (A1 before B1) and (A1 before B2). However, both instances share the same A1 element. This may not be the preferred or correct option, for example, in cases where the temporal coverage support is taken into account. Therefore we define two additional mining modes which choose only one instance by applying some decision logic. The "most recent" mode favors the closest B interval with respect to A. In our example, only (A1 before B1) is considered. The "latest" mode favors the latest B interval with respect to A. In our example, only (A1 before B2) is considered. In this study we opt to use the "most recent" mode.

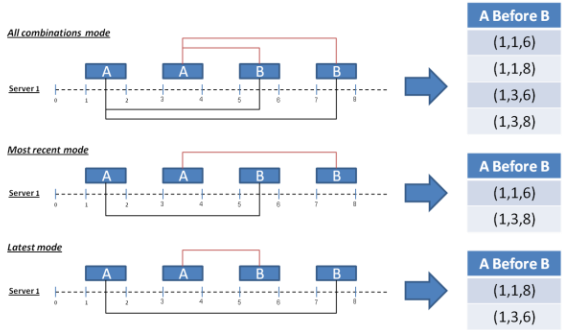

Fig. 5. Example of three possible mining modes.

## D. Pre-Processing

The algorithm presented in [11] doesn't distinct between events or facts having significantly different durations. For example, 10 seconds interval and 1 hour interval are considered the same as long as they are valid for the pattern's envelop – i.e., time window. In our implementation we perform preprocessing which descretize the duration of interval before executing the mining algorithm. Therefore, for example, each instance of the fact (BytesRecievedSec_STATE, HIGH), which refers to <u>high</u> amount of bytes received, is classified as SHORT, MEDIUM or LONG, according to predefined configuration. Consequently, given the following configuration: SHORT $\leq 60_{sec}$, $60_{sec}$ < MEDIUM $\leq 15_{min}$, LONG > $15_{min}$ A 10sec interval of (BytesRecievedSec_STATE, HIGH) will be presented as (BytesRecievedSec_STATE, HIGH, SHORT), which refers to a <u>short</u> time interval of a <u>high</u> amount of bytes received.

## I. Methodology

The proposed approach consists of the following six steps:

1. **Apply the KBTA method** on the raw data in order to derive basic state, rate and gradient abstractions.

2. **Discretize intervals** (duration) of each of the resulted abstractions from step one into VERY-SHORT, SHORT, MEDIUM, LONG and VERY-LONG.

3. **Apply the temporal pattern mining** on the discretized abstractions and identify temporal patterns that are repeated sufficiently enough to be considered as frequent; i.e., comply with the predefined horizontal support (and thus representing normal behavior).

4. **Select (manually) a sub-set of normal behavior patterns** that were identified in previous step. In this study only patterns with 2 or 3 atomic patterns were selected. These patterns are added to the domain's KBTA ontology.

5. **Apply the KBTA pattern matching mechanism on the test set** in order to identify the patterns selected in step 4.

6. **Detect abnormal time periods.** Finally, from the patterns that were identified in the test set, abnormal time periods are inferred. This is done using a *skyline* abstraction which was defined and implemented specifically for this task. The skyline abstraction is a type of state abstraction and is defined over a set of high-level abstractions (including Trend, State, Gradient, Rate and Pattern). The skyline abstraction derives time intervals and assigns a value to each time interval according to the amount of "abstracted-from" patterns that appears during that time interval.

The proposed approach of integrating the KBTA method with a temporal interval-based pattern mining algorithm is presented in Figure 6. The KBTA process is applied on raw data using a domain knowledge that is specified by the domain expert. The abstracted data is then pre-processed (e.g., indicating specific known meaningful event or discretizing the interval duration). Then, the temporal pattern mining algorithm is applied and new meaningful patterns are presented to the expert which in turn decides whether to add them to the knowledge base.

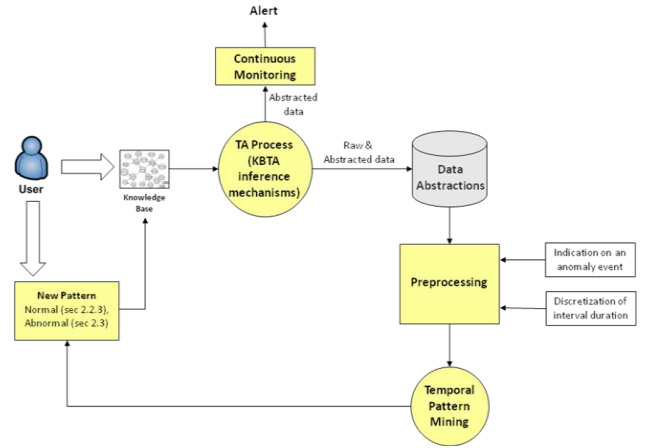

Fig. 6. The KBTA -based anomaly detection using temporal pattern mining algorithm.

## V. EVALUATION

In this section we present the preliminary evaluation that was conducted in order to demonstrate the proposed approach. In the evaluation, we attempt to extract temporal patterns which represent normal behavior of a single server. We performed a chronological experiment in which we derive normal patterns during a period of one week (only weekdays) and use the derived patterns for detecting abnormal time intervals on the following week. The mining algorithm, presented in previous section was applied on abstracted data of a server. The abstracted data was initially derived by using our KBTA framework. Next, we identify time intervals in which only few these normal patterns exist. This is done by applying the skyline abstraction. Note that in this evaluation we attempt to learn the normal behavior based on a single server's data and therefore the vertical support (i.e., the percentage of servers on which the same pattern exists) is irrelevant.

The mining process was applied on the abstracted data of a specific server, which includes the following abstractions: BytesReceivedSec_STATE, BytesReceivedSec_TREND, BytesSentSec_STATE, BytesSentSec_TREND, CPUSys_STATE, CPUSys_TREND, CPUUser_STATE, CPUUser_TREND, MemFree_STATE, MemFree_TREND, MemTotal_STATE, MemTotal_TREND, SegmentsRetransmittedSec_STATE, SegmentsRetransmittedSec_TREND, WinMemPageWritesPerS_STATE, WinMemPageWritesPerS_TREND

As presented in Section V, after the abstraction process, and before the temporal mining process, the duration of each of the abstractions was discretized into VERY-SHORT, SHORT, MEDIUM, LONG and VERY-LONG.

Figure 7 presents the Memory Free, Memory total, Bytes received\sec, CPU User, and Bytes Sent\sec state abstractions for a monitored server. When looking at the CPU-User_state parameter of the server (second graph from bottom), we can derive that the normal pattern during weekdays (Mon – Fri) is a trend of increasing from Very-Low state to Low, Medium, High and then Very-High state at midday, and then a decreasing trend from Very-High state to High, Medium, Low and finally Very-Low at the end of the day (March 22nd 2011). However, during the weekend, the normal behavioral pattern is CPU-User_state that is mostly Low or Medium.

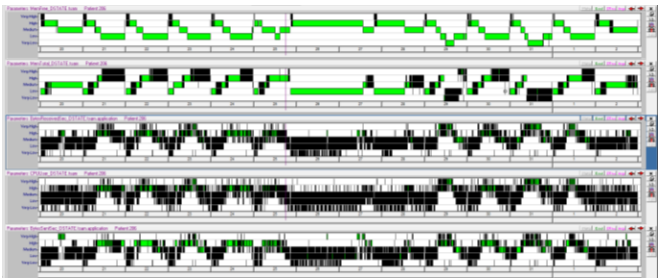

Fig. 7. Example of state abstractions for the monitored server.

Our goal is to be able to mine such normal behavioral patterns automatically in order to detect time intervals in which such patterns do not exist. As an example, we can see in Figure 6 that during the 28th (Mon) the CPU-User_state parameter does not appear as expected and actually represents more the behavior of the parameter during weekends.

### A. Evaluation Settings

The evaluation was conducted on two weeks of data (20.3.2011–02.4.2011) collected from a real server of real production environment of a large company. The first week of data (20.3.2011–26.3.2011) was used for the training. The evaluation was performed only on weekdays (i.e., 21.3.2011-25.3.2011) under the assumption that the normal behavior is different on weekdays and weekends. We attempted to derive normal (frequent patterns) that were later used for identifying time intervals in which relatively small amount of "normal" patterns exist. The settings of the mining execution are:

a. The maximum **window size** of a pattern (i.e., total duration of a mined pattern) is limited to 10 hrs.

b. **Minimum coverage support** = 3 days. This means that each mined pattern covers 3 out of 5 weekdays; i.e., at least 60% of the time.

c. **Mining algorithm mode** = most recent

Then, patterns with 2 or 3 atomic patterns were selected and added to the ontology. The KBTA pattern matching mechanism was applied to in order to identify the selected patterns in the data of the consecutive week (i.e., 28.3.2011 – 1.4.2011). Finally, a skyline abstraction was added to the ontology. The skyline abstraction is abstracted from the selected subset of patterns, and generates time intervals according to the amount of "abstracted-from" patterns that appears during that time interval.

### B. Mining Results

Applying the mining algorithm resulted in 20 patterns of size of two (i.e., two atomic patterns) and 54 patterns of size three. For example, temporal pattern 58, mined from the data of server #206, represents a long time interval (10-60min) of non-changing (i.e., SAME) Memory Total (marked as fact A) before (0-5min) a very short time interval (1min or less) of high system CPU consumption (marked as fact B) and this is during a very short time interval (1min or less) of decreasing bytes received per second (marked as fact C) (see Figure 8).

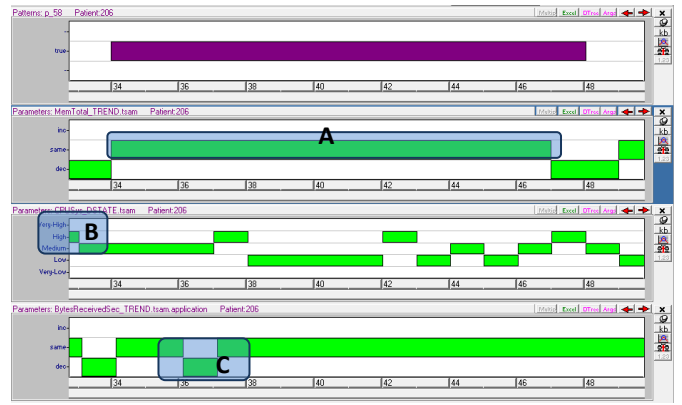

Fig. 8. Temporal pattern number 58 which was from the data of server 206.

*C. Final Evaluation Results*

In the last phase we defined a Skyline abstraction that was abstracted from a subset of the revealed patterns for each server. The Skyline abstraction categorized time intervals based on the amount of instances of the "abstracted-from" patterns that appear at each time interval. For example, if less than 10% of the "abstracted-from" patterns exist during time interval *t*, the time interval is marked as "FEW". Similarly, if 30% or more exist than the time interval is marked as "MANY". We consider time intervals which were marked as "FEW" by the skyline abstraction as potential anomalies since only few "normal" patterns exist during these time interval.

In the results we got a positive indication that the approach can be useful for detecting anomalies. From Figure 9 we identify two long time interval (approx. 13 hours) that was labeled as "Few" (i.e., in this time interval only one pattern instance at the most is observed; this indicates an abnormal period). When aligning this time interval with the primitive parameters we can visually identify correlation with the skyline abstraction: increase and then decrease of the CPU user and, Bytes Received per second and decreasing free memory.

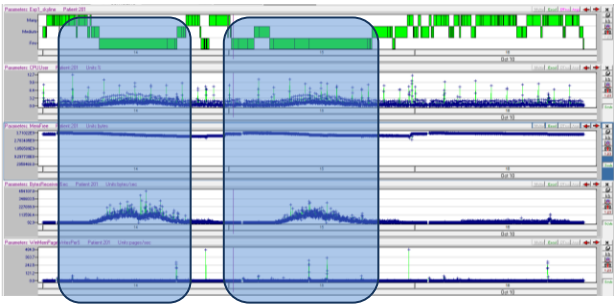

Fig. 9. The output of the Skyline abstraction in the preliminary experiment.

## VI. Conclusion and Future Work

This study presents a framework that uses the KBTA method combined with a temporal pattern mining process for anomaly detection. An experiment that was conducted on the data of a single server illustrates the process of applying the KBTA method, executing the mining process on the output abstractions and using the revealed patterns for anomaly detection. In future work we plan to extend the experiments to a larger set of data (i.e., more servers and longer period of time) and also considering additional datasets.